\documentclass[10pt,twocolumn,letterpaper]{article}

\usepackage{iccv}
\usepackage{times}
\usepackage{epsfig}
\usepackage{graphicx}
\usepackage{amsmath}
\usepackage{amssymb}

\usepackage{latexsym}
\usepackage{amsmath}
\usepackage{amssymb}
\usepackage{xcolor}
\usepackage{enumerate}
\usepackage{comment}
\usepackage{subfigure}
\usepackage{multirow}
\usepackage{bm}
\usepackage{url}

\usepackage[breaklinks=true,bookmarks=false]{hyperref}

\iccvfinalcopy 

\def\iccvPaperID{3658} 

\ificcvfinal\fi

\begin{document}

\title{ShapeCaptioner: Generative Caption Network for 3D Shapes by Learning a Mapping from Parts Detected in Multiple Views to Sentences}

\author{Zhizhong Han$^{1,3}$, Chao Chen$^{1,2}$, Yu-Shen Liu$^{1,2}$\thanks{Corresponding Author. This work was supported by National Key R\&D Program of China (2018YFB0505400) and NSF (award 1813583).}, Matthias Zwicker$^3$\\
$^1$School of Software, Tsinghua University, Beijing, China \\
$^2$Beijing National Research Center for Information Science and Technology (BNRist)\\
$^3$Department of Computer Science, University of Maryland, College Park, USA\\
{\tt\small h312h@umd.edu, thss15\_chenc@163.com, liuyushen@tsinghua.edu.cn, zwicker@cs.umd.edu}
}

\maketitle
\ificcvfinal\thispagestyle{empty}\fi

\begin{abstract}
3D shape captioning is a challenging application in 3D shape understanding. Captions from recent multi-view based methods reveal that they cannot capture part-level characteristics of 3D shapes. This leads to a lack of detailed part-level description in captions, which human tend to focus on. To resolve this issue, we propose \textit{ShapeCaptioner}, a generative caption network, to perform 3D shape captioning from semantic parts detected in multiple views. Our novelty lies in learning the knowledge of part detection in multiple views from 3D shape segmentations and transferring this knowledge to facilitate learning the mapping from 3D shapes to sentences. Specifically, ShapeCaptioner aggregates the parts detected in multiple colored views using our novel part class specific aggregation to represent a 3D shape, and then, employs a sequence to sequence model to generate the caption. Our outperforming results show that ShapeCaptioner can learn 3D shape features with more detailed part characteristics to facilitate better 3D shape captioning than previous work.
\end{abstract}

\section{Introduction}
\begin{figure}[tb]
  \centering
   \includegraphics[width=\linewidth]{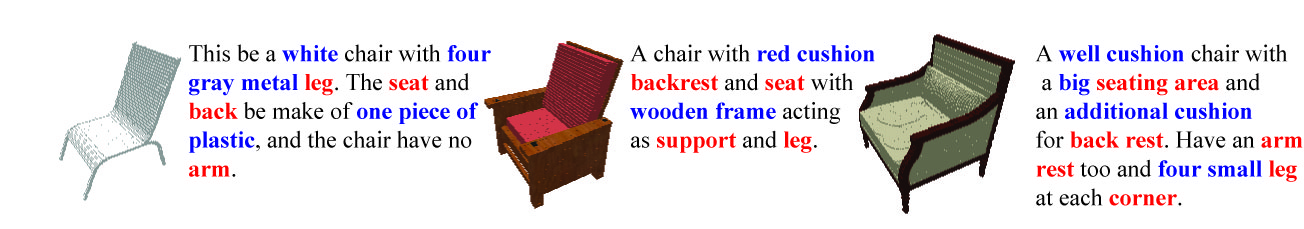}
  %
  %
\caption{\label{fig:human} Examples of human annotated captions for 3D shapes. Humans tend to describe 3D shapes by focusing on their semantic parts (in red) with their various attributes (in blue).}
\end{figure}

Jointly understanding 3D shapes and sentences is an important challenge in 3D shape analysis. For example, generated captions are helpful for visually impaired people to understand what a 3D shape looks like, including the category, color, form, and material of the 3D shape, as shown in the examples in Fig.~\ref{fig:human}. This motivates us to address the issue of automatically generating captions for 3D shapes.

Recently, Text2Shape~\cite{chenkevin2018ACCV} made an important contribution by proposing a 3D-Text dataset, where voxel-based 3D shapes and their corresponding captions are paired together. To reduce voxel complexity by representing 3D shapes as view sequences, $\rm Y^2$Seq2Seq~\cite{Zhizhong2019seq} was presented to jointly learn a bilateral mapping between view sequences and word sequences. Although $\rm Y^2$Seq2Seq can produce plausible captions for 3D shapes, its ability to generate detailed description for parts is limited because the model does not capture part-level characteristics. However, human-provided captions often focus on part-level details, as the manually annotated ground truth examples in Fig.~\ref{fig:human} illustrate. Therefore, how to absorb part characteristics in 3D shape captioning remains challenging.

To resolve this issue, we propose a novel deep neural network, \textit{ShapeCaptioner}, to generate captions for 3D shapes. By representing a 3D shape as a view sequence, ShapeCaptioner aims to learn a mapping from parts detected in the view sequence to a caption describing the 3D shape. Specifically, ShapeCaptioner first leverages 3D shape segmentation benchmarks to learn the knowledge of detecting parts in terms of their geometry from multiple views, and then, transfers this knowledge to incrementally learn the ability of detecting parts with various attributes from multiple colored views in 3D-Text dataset. ShapeCaptioner further employs a novel part class specific aggregation to aggregate the detected parts over all views for the captioning of a 3D shape.
The part class specific aggregation can represent a 3D shape by capturing more part characteristics from different views, which facilitates more detailed 3D shape captioning.
In summary, our contributions are as follows:

\begin{enumerate}[i)]
\item We propose ShapeCaptioner to enable 3D shape captioning from semantic parts detected in multi-views, which facilitates more detailed 3D shape captioning.
\item We introduce a method to learn the knowledge of part detection in multi-views from 3D shape segmentation benchmarks to facilitate 3D shape captioning.
\item With our novel part class specific aggregation, we effectively capture part charicteristics to represent 3D shapes for better shape captioning, inspired by the way humans describe 3D shapes in terms of semantic parts.
\end{enumerate}

\section{Related work}
\noindent\textbf{Image captioning.} There is a large volume of work on image captioning. Here we review methods based on object detection, which are most similar to our work. Currently, end-to-end deep learning approaches~\cite{KarpathyL15,Karpathy:2017} are most effective for image captioning. These methods try to learn to generate captions from global image features. However, using a global feature limits their interpretability. Other work~\cite{SCN_CVPR2017,wu7780398,YouJWFL16,YinO17emnlp,WangMS18,densecap16} employs object-level semantics to generate captions. These methods represent images based on occurring semantic concepts or objects. In~\cite{YinO17emnlp}, explicitly detected objects are employed with their category, size and layout to generate captions. In contrast, based on a bag of word model, impressive captions can also be generated by only using the explicitly detected objects~\cite{WangMS18}.

Our method is different from these methods in four aspects. First, the parts we want to detect are more various than the objects. Second, part detection becomes more challenging when considering multiple unaligned and varying viewpoints. Third, how to aggregate the detected parts over different views to enable learning the mapping from 3D shapes to captions represents an additional problem. Fourth, a final obstacle is that there is no labeled dataset available to learn the knowledge of part detection in multiple views of 3D shapes. Yet ShapeCaptioner can resolve these issues to generate captions with detailed part characteristics.

\noindent\textbf{3D shape captioning.} Although deep learning models have led to significant progress in feature learning for 3D shapes~\cite{Zhizhong2016b,Zhizhong2016,Han2017,HanTIP18,Zhizhong2018seq,MAPVAE19,parts4features19,3DViewGraph19,3D2SeqViews19,HanCyber17a,Zhizhong2018VIP,p2seq18}, 3D shape captioning has been less explored due to the lack of training dataset. However, the recently proposed 3D-Text dataset~\cite{chenkevin2018ACCV} has enabled the research in this area. $\rm Y^2$Seq2Seq~\cite{Zhizhong2019seq} employs multiple views as a 3D shape representation to address the cubic complexity of voxel representations. It learns 3D shape features by aggregating the global feature of each view. Although $\rm Y^2$Seq2Seq can generate plausible captions for 3D shapes, it often fails to generate captions with local part details. To resolve this issue, ShapeCaptioner represents a 3D shape as a set of parts detected in multiple views, which not only avoids the cubic complexity of voxels but also enables the ability of capturing part characteristics. This leads to captions that are more similar to the manually annotated ground truth, which usually includes part details, such as color, form, material, and texture of parts shown in Fig.~\ref{fig:human}.

\section{Overview}
\begin{figure*}[tb]
  \centering
   \includegraphics[width=\linewidth]{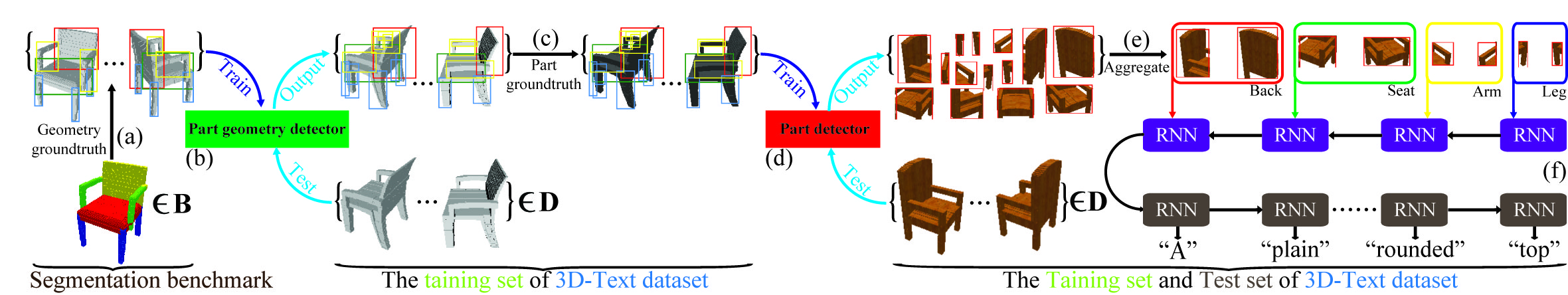}
  %
  %
\caption{\label{fig:framework} The demonstration of ShapeCaptioner. ShapeCaptioner first learns the knowledge of detecting part geometry in multiple views from shape segmentation benchmark $\mathbf{B}$ ((a) and (b)). By transferring this knowledge to 3D-Text dataset $\mathbf{D}$, ShapeCaptioner learns the ability of detecting parts containing both geometry and attributes ((b), (c) and (d)). Finally, ShapeCaptioner generates captions by learning the mapping from the detected parts to the description ((d), (e) and (f)).}
\end{figure*}

ShapeCaptioner learns to generate a caption $\bm{t}$ for a 3D shape $\bm{s}$ under a 3D-Text dataset $\mathbf{D}$ which provides ground truth $(\bm{s},\bm{t})$ pairs, where shape $\bm{s}$ is represented by a $128^3$ dimensional voxel cube and contains various attributes, such as color, material and texture, as demonstrated in Fig.~\ref{fig:human}. ShapeCaptioner represents shape $\bm{s}$ as a colored view sequence $\bm{v}=\{v_i, i\in[1,V]\}$, and detects important parts from each view $v_i$ to capture part characteristics for caption generation. However, dataset $\mathbf{D}$ does not have ground truth parts in multiple views for ShapeCaptioner to learn from.

To resolve this problem, we first leverage a separate 3D shape segmentation benchmark $\mathbf{B}$ to learn the knowledge of part geometry detection in multiple views, i.e., detecting parts only in terms of geometry rather than attributes, as demonstrated in Fig.~\ref{fig:framework} (a) and (b). This is because there is no color, material or texture available in benchmark $\mathbf{B}$. Then, we transfer this knowledge to 3D-Text dataset $\mathbf{D}$ to establish the ground truth parts in multiple colored views which enables ShapeCaptioner to incrementally learn the ability of detecting parts in terms of both geometry and attributes, as demonstrated in Fig.~\ref{fig:framework} (b), (c) and (d). Finally, ShapeCaptioner learns a mapping from parts detected in multiple views of shape $\bm{s}$ to its description $\bm{t}$ to facilitate 3D shape captioning, as demonstrated in Fig.~\ref{fig:framework} (d), (e) and (f).

\section{ShapeCaptioner}
\noindent\textbf{Part geometry detection. }As shown in Fig.~\ref{fig:framework} (a), we propose a method to obtain part geometry ground truth in multiple views from 3D shape segmentation benchmark $\mathbf{B}$. We use the 3D shape segmentation benchmarks involved in~\cite{KalogerakisAMC17} as segmentation ground truth.

\begin{figure}[tb]
  \centering
   \includegraphics[width=\linewidth]{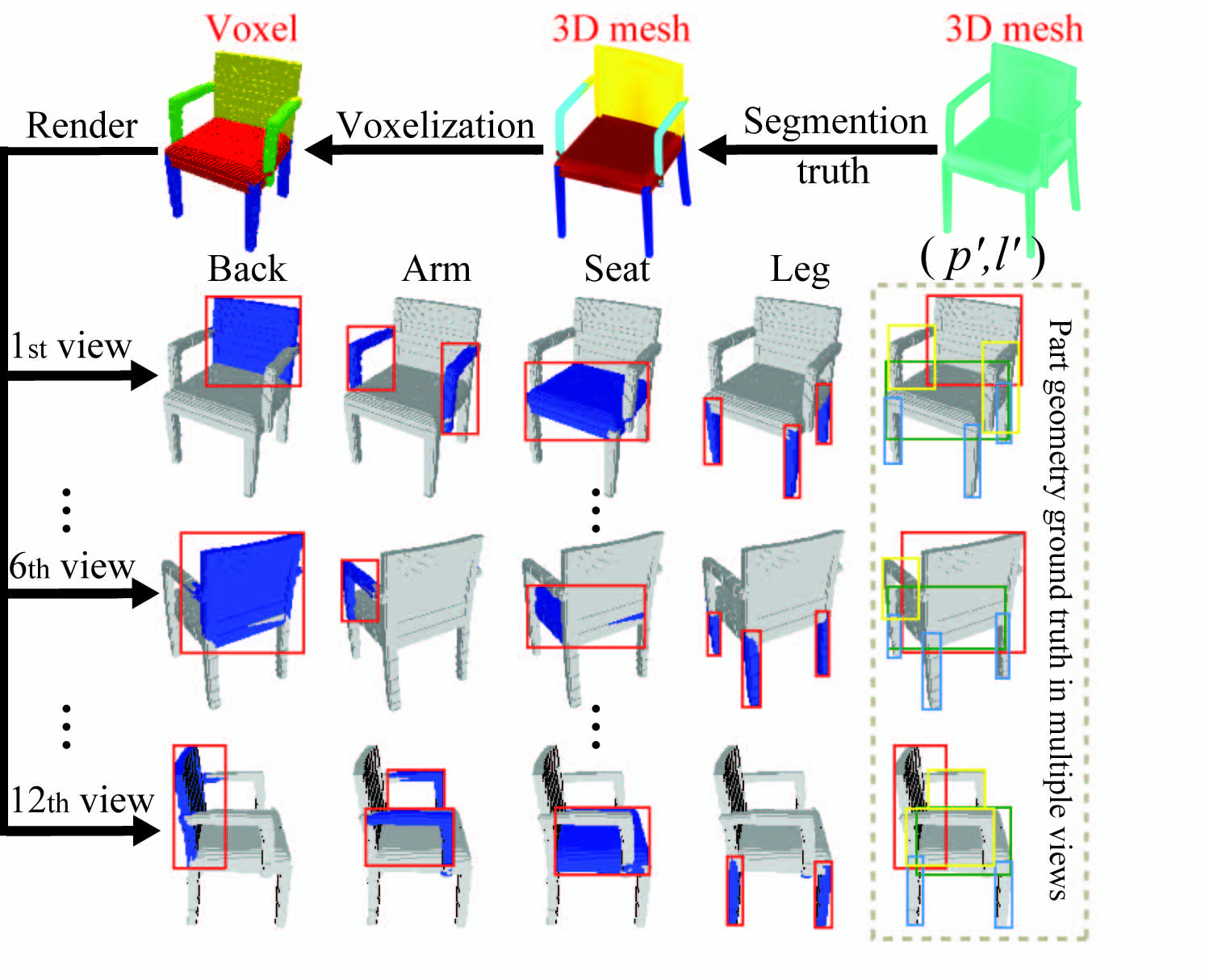}
  %
  %
\caption{\label{fig:partGT} Illustration of generating part geometry ground truth in multiple views from 3D shape segmentation benchmarks.}
\end{figure}

As illustrated in Fig.~\ref{fig:partGT}, starting from each 3D mesh in benchmak $\mathbf{B}$, our method first voxelizes the 3D mesh into voxels, along with labelling each voxel according to the segmentation ground truth on the mesh. In the voxelization, we randomly sample 100 points on each triangle face of the mesh, and label each sampled point by the label of the triangle face. This enables us to perform label voting among points located in the same voxel in the labelling of each voxel. Using randomly sampled points could resist the imbalance effect of triangle face size. Then, we render the voxelized shape from 12 viewpoints and locate different part geometries in each view. Specifically, from each viewpoint, we separately highlight voxels belonging to the same part class in blue, and compute the bounding box of each blue region. We denote $p'$ as a one hot probability distribution to indicate which part class the bounding box belongs to, and denote $l'$ as the location of the bounding box, where $(p',l')$ forms the part geometry ground truth.

Take the first viewpoint in Fig.~\ref{fig:partGT} for example, we separately obtain the bounding box of the region formed by voxels in back, arm, seat and leg class of a chair, and finally, we obtain the part geometry ground truth in the first view. By repeating this process, we obtain the part geometry ground truth $(p',l')$ in multiple views, as shown by the bounding box on shapes in the dashed box in Fig.~\ref{fig:partGT}.

The reason why we obtain views with part geometry ground truth by rendering the voxelized shapes in benchmark $\mathbf{B}$ is to avoid the domain gap. Because 3D shapes in 3D-Text dataset $\mathbf{D}$ are voxelized, there would be a domain gap if we learn from part geometry ground truth in views rendered from meshes, while detecting part geometry from views rendered from these voxelized shapes.

Subsequently, we train a fasterRCNN~\cite{Ren2015} as a part geometry detector $G$ under the obtained part geometry ground truth $(p',l')$, as shown in Fig.~\ref{fig:framework} (b). This enables ShapeCaptioner to detect part geometries $(p_g,l_g)$ from any view $v$ by minimizing the objective function below,

\begin{equation}
\label{eq:geometry}
O_{G}(p_g,p',l_g,l')=O_{p}(p_g,p')+\lambda O_{l}(l_g,l'),
\end{equation}

\noindent where $O_{p}$ measures the accuracy in terms of probability by the cross-entropy function of part class labels, while $O_{l}$ measures the accuracy in terms of location by the robust $L_1$ function as in~\cite{Girshick2015}. The parameter $\lambda$ balances $O_{p}$ and $O_{l}$, and a value of 1 works well in all our experiments.

\noindent\textbf{Parts detection. }Part geometry detection is not enough for ShapeCaptioner to generate captions with detailed part characteristics, since the detected geometries lack various attributes, such as color or texture. Hence, it also needs to be able to detect parts with attributes.

To resolve this issue, we transfer the learned knowledge of part geometry detection to the 3D-Text dataset $\mathbf{D}$. We introduce a bounding box mapping method to establish part ground truth which contains both geometry and attributes for ShapeCaptioner to learn from.

\begin{figure}[tb]
  \centering
   \includegraphics[width=\linewidth]{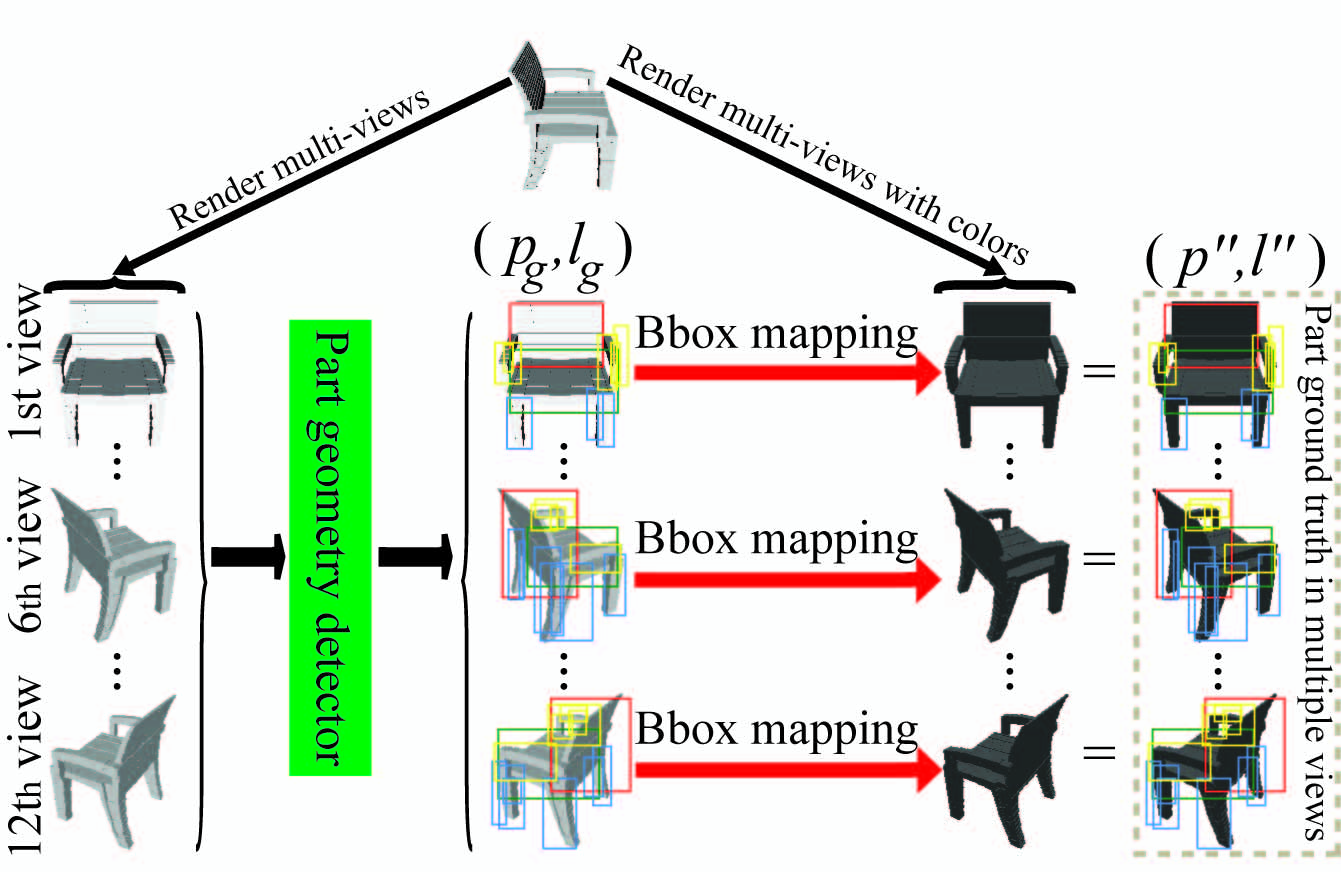}
  %
  %
\caption{\label{fig:part} Illustration of generating part ground truth in multiple views from 3D-Text dataset.}
\end{figure}

As demonstrated in Fig.~\ref{fig:framework} (b), in the test stage of part geometry detector $G$, we first render multiple views from each 3D shape in the training set of dataset $\mathbf{D}$ without color, and then, employ the trained part geometry detector $G$ to detect part geometries $(p_g,l_g)$ in these views. This process is further illustrated by the first column in Fig.~\ref{fig:part}. Subsequently, we render the 3D shape again but with color, and map the detected part geometries $(p_g,l_g)$ to the corresponding views with color. Finally, we regard the mapped part geometries in views with color as the part ground truth $(p'',l'')$, where $l''=l_g$ and $p''$ is a one hot probability distribution by setting the entry of $argmax(p_g)$ to be 1. Here, we map the detected part geometries with $p_g>0.7$.

Finally, we further push the part geometry detector $G$ to detect parts $(p,l)$ from multiple views with color by fine-tuning $G$ under the part ground truth $(p'',l'')$. We rename $G$ as part detector $R$, as shown in Fig.~\ref{fig:framework} (d). Note that these part ground truth $(p'',l'')$ are only obtained from shapes in the training set of 3D-Text dataset $\mathbf{D}$. After training, $R$ can detect parts from multiple colored views of shapes in the test set of dataset $\mathbf{D}$. Similar to Eq.~(\ref{eq:geometry}), we minimize the following objective function with a $\lambda$ of 1,

\begin{equation}
\label{eq:parts}
O_{R}(p,p'',l,l'')=O_{p}(p,p'')+\lambda O_{l}(l,l'').
\end{equation}

\noindent\textbf{Part class specific aggregation. }Given a 3D shape $\bm{s}$, ShapeCaptioner generates its caption $\bm{t}$ from the parts detected in multiple colored views $v_i$ of $\bm{s}$, as demonstrated in Fig.~\ref{fig:framework} (e). We denote the $j$-th part detected by $R$ from $v_i$ as $(p_i^j,l_i^j)$. We select the detected parts with $p_i^j>\rho$ to represent the 3D shape $\bm{s}$, where $\rho$ is a probability threshold.

ShapeCaptioner aggregates the selected parts $\{(p_i^j,l_i^j)\}$ over all $V$ views in terms of different part classes to represent shape $\bm{s}$, as demonstrated in Fig.~\ref{fig:framework} (e). For each one of $C$ part classes, we group the parts $(p_i^j,l_i^j)$ belonging to the same $c$-th part class if $argmax(p_i^j)$ is $c$. Then, we obtain a part class specific feature $\bm{F}_c$ by aggregating all parts in the $c$-th part class using pooling procedure as follows,

\begin{equation}
\label{eq:pool}
\bm{F}_c=\text{pool}_{argmax(p_i^j)==c} (\bm{f}_i^j), i\in[1,V],
\end{equation}

\noindent where $\bm{f}_i^j$ is a 4096 dimensional feature extracted from the fc7 layer of part detector $R$. Finally, we use $\bm{F}$ to represent shape $\bm{s}$ as a sequence of part class specific feature $\bm{F}_c$ as follows, where $c\in[1,C]$,

\begin{equation}
\label{eq:pool}
\bm{F}=[\bm{F}_1,...,\bm{F}_c,...,\bm{F}_C].
\end{equation}

The reason we represent a 3D shape $\bm{s}$ as a combination of part class specific features $\bm{F}_c$ is to preserve as much part characteristics as possible in the aggregation process. This could reduce the impact of one part class on the others in the multi-view scenario, which enables $\bm{F}_c$ to comprehensively describe what the semantic part looks like over all $V$ views of shape $\bm{s}$.


\noindent\textbf{Captioning from parts. }ShapeCaptioner leverages the sequence $\bm{F}$ of part class specific features $\bm{F}_c$ from $\bm{s}$ to generate a sequence of words $t_n$ as a caption $\bm{t}$, where $\bm{t}=[t_1,...,t_n,...,t_N]$ and $n\in [1,N]$. We cast this problem into a sequence to sequence translation model, and implement this seq2seq model by a RNN encoder and a RNN decoder, as demonstrated in Fig.~\ref{fig:framework} (f). The RNN encoder encodes $\bm{F}$ by inputting $\bm{F}_c$ at each one of $C$ steps, while the RNN decoder dynamically decodes each word $t_n$ in $\bm{t}$. Thus, ShapeCaptioner learns the mapping from 3D shape to sentences by minimizing the following objective function,

\begin{equation}
\label{eq:caption}
O_{\bm{t}}=-\sum_{t_n\in\bm{t}}\log p(t_n|t_{<n},\bm{F}),
\end{equation}

\noindent where $t_n$ is the $n$-th word in the word sequence $\bm{t}$, $t_{<n}$ represents the words in front of $t_n$, $p(t_n|t_{<n},\bm{F})$ is the probability of correctly predicting the $n$-th word according to the previous words $t_{<n}$ and the 3D shape feature $\bm{F}$. Note that the optimization is conducted under the training set of 3D-Text dataset $\mathbf{D}$. After training, we can generate captions for shapes in the test set of dataset $\mathbf{D}$.

%

\section{Experimental results and analysis}
We evaluate ShapeCaptioner in this section. We first explore the effects of some important parameters on the performance of ShapeCaptioner. Then, we conduct some ablation studies to justify the effectiveness of some modules. Finally, we compare it with some state-of-the-art methods.

\noindent\textbf{Dataset and metrics. }We evaluate ShapeCaptioner under the 3D-Text cross-modal dataset~\cite{chenkevin2018ACCV}, which consists of a primitive subset and a ShapeNet subset. We only employ the ShapeNet subset, because the 3D shapes in the primitive subset are too simple to extract parts. The ShapeNet subset contains 15,038 shapes and 75,344 descriptions in the chair and table classes. We employ the same training/test splitting in each shape class as~\cite{chenkevin2018ACCV,Zhizhong2019seq}. Specifically, the chair class is formed by 5954 training shapes and 641 test shapes while the table class contains 7592 training shapes and 851 test shapes. In addition, we employ 487 chairs (out of 537) and 481 tables (out of 520) involved in segmentation benchmarks~\cite{KalogerakisAMC17} to train part geometry detector $G$, respectively, which avoids the 3D shapes that are also in the test set of 3D-Text dataset.

We employ BLUE~\cite{Papineni2002BMA}, CIDEr~\cite{VedantamZP15}, METEOR~\cite{W14-3348}, and ROUGE~\cite{Lin:2004} to evaluate the quality of generated captions according to the ground truth captions, where these metrics are abbreviated as ``B-1'', ``B-2'', ``B-3'', ``B-4'', and ``C'', ``M'', ``R'', respectively in the following tables.

\begin{figure*}[tb]
  \centering
   \includegraphics[width=\linewidth]{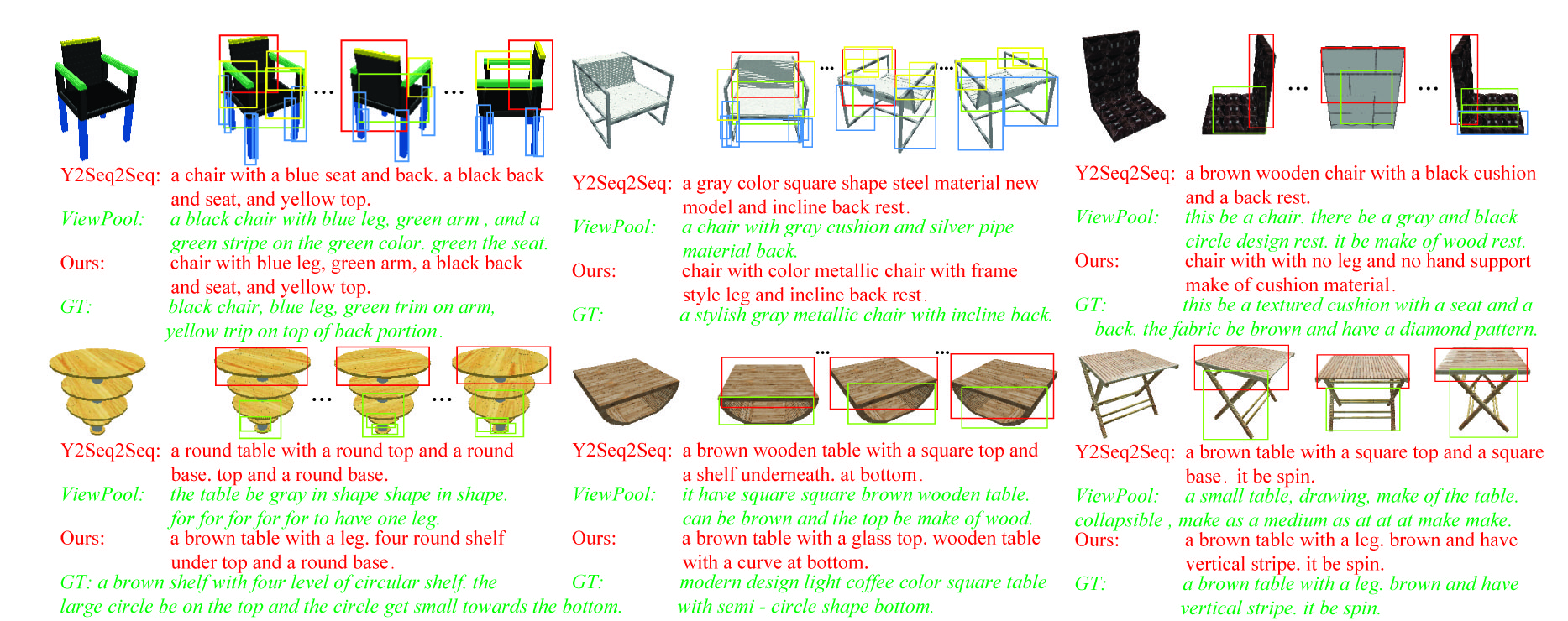}
  %
  %
\caption{\label{fig:CapCompare} Comparison between captions generated by different methods. Compared to ground truth, our method can generate captions with more accurate part characteristics, where $\rho>0.8$.}
\end{figure*}

\noindent\textbf{Initialization. }We extract 3587 unique words from captions of 3D shapes in the training set of ShapeNet subset to form the vocabulary for caption generation. Each word is represented by a 512 dimensional embedding, which is learned along with the other involved parameters in training. Both RNN encoder and decoder are implemented by GRU cells~\cite{Cho2014On}. The learning rate in all the experiments is 0.00001.

In addition, the number of part classes is $C=4$ in the chair class, while $C=3$ in the table class, and ShapeCaptioner is trained under each class respectively. We employ $\rho=0.8$ to select parts detected in $V=12$ views to represent a 3D shape, and then, use max pooling to aggregate all the selected parts over $V$ views.


\noindent\textbf{Parameters. }Here we compare some important parameters under the chair class.

\begin{figure}[tb]
  \centering
   \includegraphics[width=\linewidth]{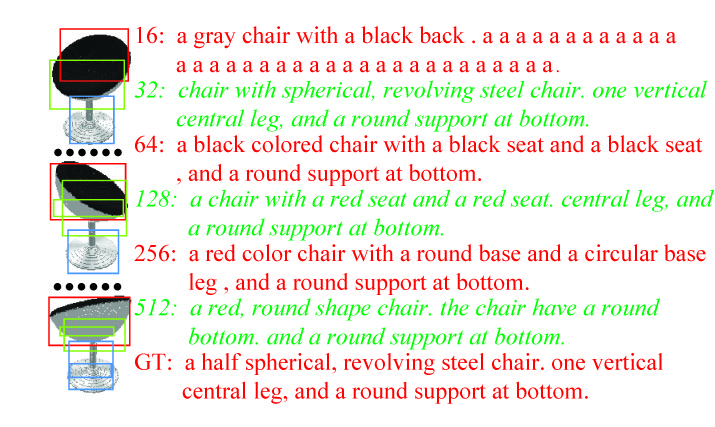}
  %
  %
\caption{\label{fig:hidden} Comparison between generated captions under different dimensions of the RNN hidden state.}
\end{figure}

We first explore the effect of dimension $H$ of the RNN encoder and decoder by comparing $H\in\{16,32,64,128,256,512\}$. As shown in Table~\ref{table:metricH}, the performance is increased with increasing $H$ until $H=32$, and then, degenerates gradually when $H$ becomes larger. We believe this is caused by overfitting because the training data is not large enough. In addition, we compare the generated captions under different $H$ in Fig.~\ref{fig:hidden}, where the detected parts for caption generation are also briefly shown on the views. We also observe the gradually degenerated captions when $H$ becomes larger. For example, the color description becomes inaccurate in the caption with $H=128$. Also, some descriptions are repeating in the captions with $H=256$ and $H=512$. In the following, we set $H$ to 32.

\begin{table}
  \caption{The comparison on $H$. $\rho=0.8$, $V=12$.}%
  \label{table:metricH}
  \centering
  \resizebox{0.5\textwidth}{!}{
  \begin{tabular}{c|c|c|c|c|c|c|c}
    \hline
    $H$&B-1&B-2&B-3&B-4&M&R&C\\
    \hline
    16&0.122&0.072&0.041&0.021&0.121&0.168&0.002\\
    32&\textbf{0.937}&\textbf{0.917}&\textbf{0.894}&\textbf{0.878}&\textbf{0.550}&\textbf{0.847}&\textbf{1.789}\\
    64&0.794&0.639&0.537&0.475&0.284&0.551&0.644\\
    128&0.761&0.604&0.493&0.421&0.273&0.529&0.560\\
    256&0.772&0.615&0.503&0.431&0.276&0.534&0.576\\
    512&0.736&0.559&0.442&0.371&0.265&0.503&0.506\\
    \hline
  \end{tabular}}
\end{table}

\begin{figure}[tb]
  \centering
   \includegraphics[width=\linewidth]{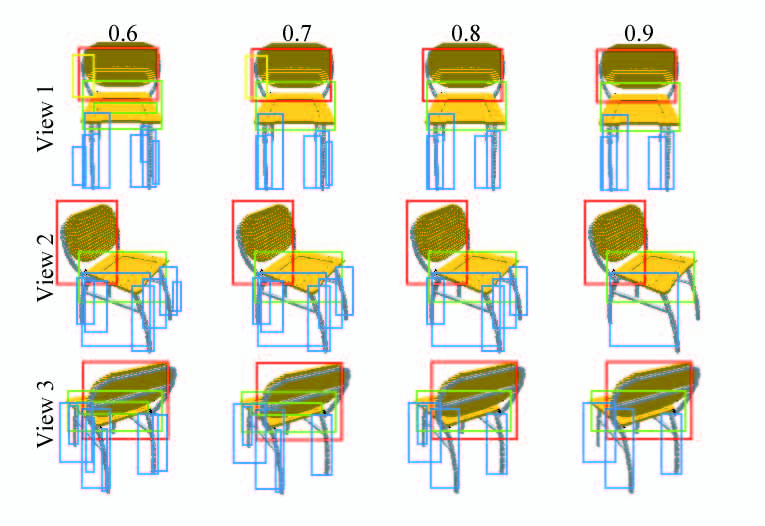}
  %
  %
\caption{\label{fig:rho} Comparison between detected parts in three views under different probability thresholds $\rho$.}
\end{figure}

Then, we explore the probability threshold $\rho$ of selecting parts to represent a 3D shape. We compare $\rho\in\{0.6,0.7,0.8,0.9\}$. There would be more selected parts when $\rho$ is smaller while the quality of the selected parts is lower, and vice versa, as demonstrated in Fig.~\ref{fig:rho}. We believe both the number and the quality would affect the discriminability of 3D shape features. This is because smaller number of parts would decrease the ability of resisting the effect of viewpoint changing while lower quality would contain inaccurate part characteristics. In Table~\ref{table:metricRho}, we found $\rho=0.8$ performs best, hence we use this setting in the following.

\begin{table}
  \caption{The comparison on $\rho$. $H=32$, $V=12$.}
  \label{table:metricRho}
  \centering
  \resizebox{0.5\textwidth}{!}{
  \begin{tabular}{c|c|c|c|c|c|c|c}
    \hline
    $\rho$&B-1&B-2&B-3&B-4&M&R&C\\
    \hline
    0.6&0.225&0.140&0.085&0.053&0.154&0.261&0.046\\
    0.7&0.715&0.576&0.476&0.415&0.260&0.515&0.523\\
    0.8&\textbf{0.937}&\textbf{0.917}&\textbf{0.894}&\textbf{0.878}&\textbf{0.550}&\textbf{0.847}&\textbf{1.789}\\
    0.9&0.463&0.409&0.383&0.366&0.235&0.366&0.428\\
    \hline
  \end{tabular}}
\end{table}

\begin{figure}[tb]
  \centering
   \includegraphics[width=\linewidth]{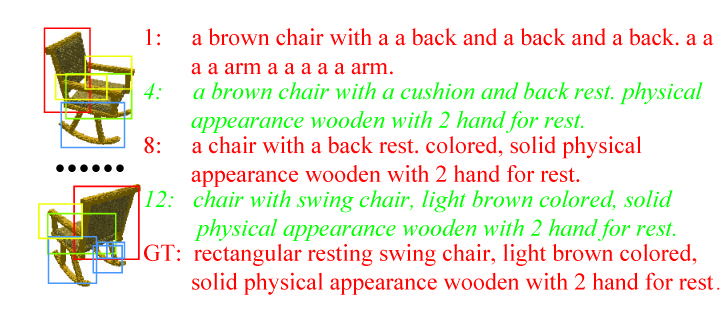}
  %
  %
\caption{\label{fig:viewnum} Comparison between generated captions under different numbers of views.}
\end{figure}

Finally, we want to know how the number $V$ of views affects the performance. In this experiment, we use the first $\{1,4,8,12\}$ views to select parts, and then, learn to generate captions. As shown in Table~\ref{table:metricV}, the performance keeps improving along with the increasing number of views. This comparison shows that more views would provide more part characteristics to learn from, which also decreases the effect of inaccurately detected parts. This can also be observed in the comparison of the generated captions under different $V$ in Fig.~\ref{fig:viewnum}, where the parts employed for caption generation are also briefly shown. For example, we cannot get plausible captions when too few views are available, such as $V=1$. The material and the color gradually appear in the captions when $V$ increases from 4 to 12. We do not explore the results with more views because of the limited computational capacity. In the following, we use $V=12$.

\begin{table}
  \caption{The comparison on $V$. $H=32$, $\rho=0.8$.}
  \label{table:metricV}
  \centering
  \resizebox{0.5\textwidth}{!}{
  \begin{tabular}{c|c|c|c|c|c|c|c}
    \hline
    $V$&B-1&B-2&B-3&B-4&M&R&C\\
    \hline
    1&0.324&0.210&0.137&0.097&0.158&0.320&0.061\\
    4&0.777&0.638&0.540&0.478&0.292&0.567&0.724\\
    8&0.876&0.779&0.707&0.656&0.374&0.674&1.108\\
    12&\textbf{0.937}&\textbf{0.917}&\textbf{0.894}&\textbf{0.878}&\textbf{0.550}&\textbf{0.847}&\textbf{1.789}\\
    \hline
  \end{tabular}}
\end{table}

\begin{figure}[tb]
  \centering
   \includegraphics[width=\linewidth]{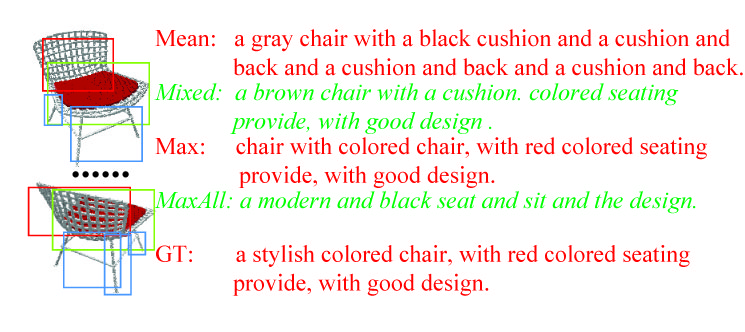}
  %
  %
\caption{\label{fig:pooling} Comparison between generated captions under different part aggregation methods.}
\end{figure}

\noindent\textbf{Ablation studies. }We first highlight our part class specific aggregation for representing a 3D shape. In the former experiments, we employ max pooling to obtain the part class specific aggregation $\bm{F}_c$. Here, we try mean pooling to compare with the results of max pooling. As shown by the result of ``Mean'' in Table.~\ref{table:pooling}, we find that mean pooling is not as good as max pooling (``Max'') to aggregate parts in multiple views. To further justify this point, we conduct another experiment to combine max pooling and mean pooling together. Specifically, we use max pooling to aggregate parts in the same part class in the same view while further using mean pooling to obtain $\bm{F}_c$ by aggregating the same part class over different views. As shown by the result of ``Mixed'', although it is better than the result of ``Mean'', it is still worse than the result of ``Max''. In addition, we also highlight the idea of part class specific features. As shown by the result of ``MaxAll'', we max pool all parts over views into a single feature while ignoring the part class, and leverage this feature to generate captions. The degenerated results show that parts in different classes would affect each other in the aggregation. In addition, we compare our employed GRU with LSTM cell in RNN. As shown by ``Max(L)'', GRU cell is more suitable in our problem. The captions in Fig.~\ref{fig:pooling} also show the similar comparison results.

\begin{table}
  \caption{The part aggregation. $H=32$, $\rho=0.8$, $V=12$.}
  \label{table:pooling}
  \centering
  \resizebox{0.5\textwidth}{!}{
  \begin{tabular}{c|c|c|c|c|c|c|c}
    \hline
    &B-1&B-2&B-3&B-4&M&R&C\\
    \hline
    Mean&0.220&0.133&0.077&0.044&0.148&0.253&0.018\\
    Mixed&0.843&0.744&0.668&0.617&0.360&0.653&1.019\\
    Max(L)&0.722&0.541&0.433&0.374&0.239&0.482&0.422\\
    MaxAll&0.714&0.593&0.520&0.476&0.299&0.535&0.694\\
    Max&\textbf{0.937}&\textbf{0.917}&\textbf{0.894}&\textbf{0.878}&\textbf{0.550}&\textbf{0.847}&\textbf{1.789}\\
    \hline
  \end{tabular}}
\end{table}

Subsequently, we highlight the advantage of our part based features over view based features. Similar to~\cite{Zhizhong2019seq}, we employ a VGG19~\cite{Simonyan14c} to extract the feature of each view, and employ a RNN encoder to aggregate these view features for caption generation. As shown in Table.~\ref{table:part}, without the multi-task scenario and constraints in~\cite{Zhizhong2019seq}, this aggregation by RNN (``RNN'') cannot obtain satisfactory results. In addition, we also try a CNN+RNN architecture similar to the method of image captioning~\cite{Vinyals43274}. We employ the first view of each shape to caption a 3D shape. Although the results of ``CNN'' are a little bit better than ``RNN'', there is still room to improve. We further try to pool the $V=12$ view features together into a single feature, and convey this feature to the RNN decoder for caption generation. We find both mean pooling (``VMean'') and max pooling (``VMax'') works well on aggregating views to generate captions, and mean pooling is better than max pooling. However, all these view based shape features cannot capture part characteristics to generate better captions than our part based features (``Part''). Moreover, we also observe similar results in the table class in Table~\ref{table:part1}.

\begin{figure*}[tb]
  \centering
   \includegraphics[width=\linewidth]{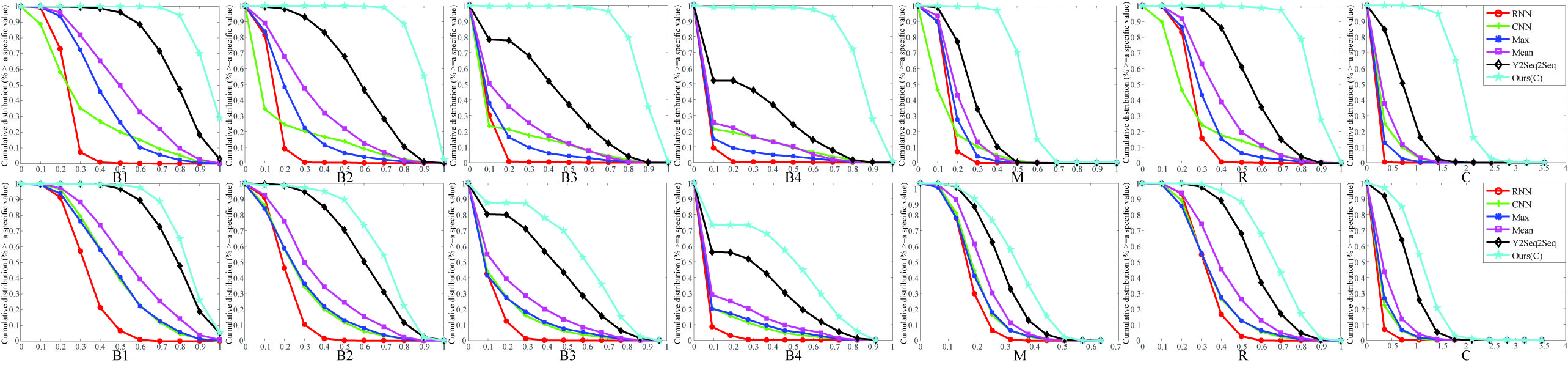}
  %
  %
\caption{\label{fig:CapCompareVis} The comparison with methods in Table~\ref{table:part}, Table~\ref{table:part1} and state-of-the-art under the test set of chair (1st row) and table (2nd row) class. The comparison is conducted in terms of cumulative distribution of different metrics.}
\end{figure*}

In addition, we elaborate on the results in Table~\ref{table:part} and Table~\ref{table:part1} by cumulative distribution in Fig.~\ref{fig:CapCompareVis}. For each metric, we use 11 values as probes, and calculate the percentage of samples over the whole test set whose metric scores are bigger than each probe, respectively. These comparisons also demonstrate our significant improvement by higher percentage.

\begin{table}
  \caption{Representations in chair class.$H=32$,$\rho=0.8$,$V=12$.}
  \label{table:part}
  \centering
  \resizebox{0.5\textwidth}{!}{
  \begin{tabular}{c|c|c|c|c|c|c|c}
    \hline
    &B-1&B-2&B-3&B-4&M&R&C\\
    \hline
    RNN&0.231&0.148&0.090&0.055&0.172&0.252&0.009\\
    CNN&0.337&0.218&0.173&0.152&0.153&0.285&0.221\\
    VMax&0.409&0.244&0.181&0.157&0.176&0.307&0.229\\
    VMean&0.494&0.338&0.251&0.214&0.209&0.381&0.301\\
    Part&\textbf{0.937}&\textbf{0.917}&\textbf{0.894}&\textbf{0.878}&\textbf{0.550}&\textbf{0.847}&\textbf{1.789}\\
    \hline
  \end{tabular}}
\end{table}

\begin{table}
  \caption{Representations in table class.$H=32$,$\rho=0.8$,$V=12$.}
  \label{table:part1}
  \centering
  \resizebox{0.5\textwidth}{!}{
  \begin{tabular}{c|c|c|c|c|c|c|c}
    \hline
    &B-1&B-2&B-3&B-4&M&R&C\\
    \hline
    RNN&0.310&0.194&0.110&0.054&0.161&0.317&0.073\\
    CNN&0.451&0.279&0.186&0.143&0.190&0.343&0.228\\
    VMax&0.450&0.285&0.210&0.180&0.192&0.344&0.283\\
    VMean&0.530&0.367&0.274&0.229&0.225&0.414&0.385\\
    Part&\textbf{0.860}&\textbf{0.755}&\textbf{0.675}&\textbf{0.620}&\textbf{0.362}&\textbf{0.664}&\textbf{1.099}\\
    \hline
  \end{tabular}}
\end{table}

\begin{figure*}[tb]
  \centering
   \includegraphics[width=\linewidth]{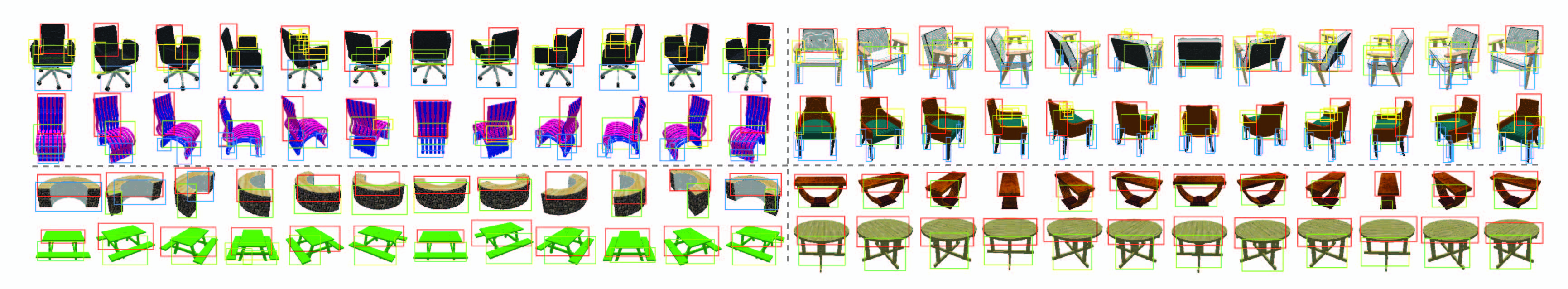}
  %
  %
\caption{\label{fig:PartVis} Demonstration of part detection under 3D-Text dataset, where $\rho>0.8$. The same color in the first two rows (four chairs) or the last two rows (four tables) indicates the same part class.}
\end{figure*}

\noindent\textbf{Visualization of detected parts.} ShapeCaptioner employs an effective way of detecting parts from multiple colored views of a 3D shape. As demonstrated by the consistent part detection results in Fig.~\ref{fig:PartVis}, where the bounding boxes of the detected parts are shown on all the $V=12$ views, ShapeCaptioner can detect reasonable parts for caption generation by understanding the complex geometry of semantic parts without the impacts by viewpoints and colors.

\begin{figure*}[tb]
  \centering
   \includegraphics[width=\linewidth]{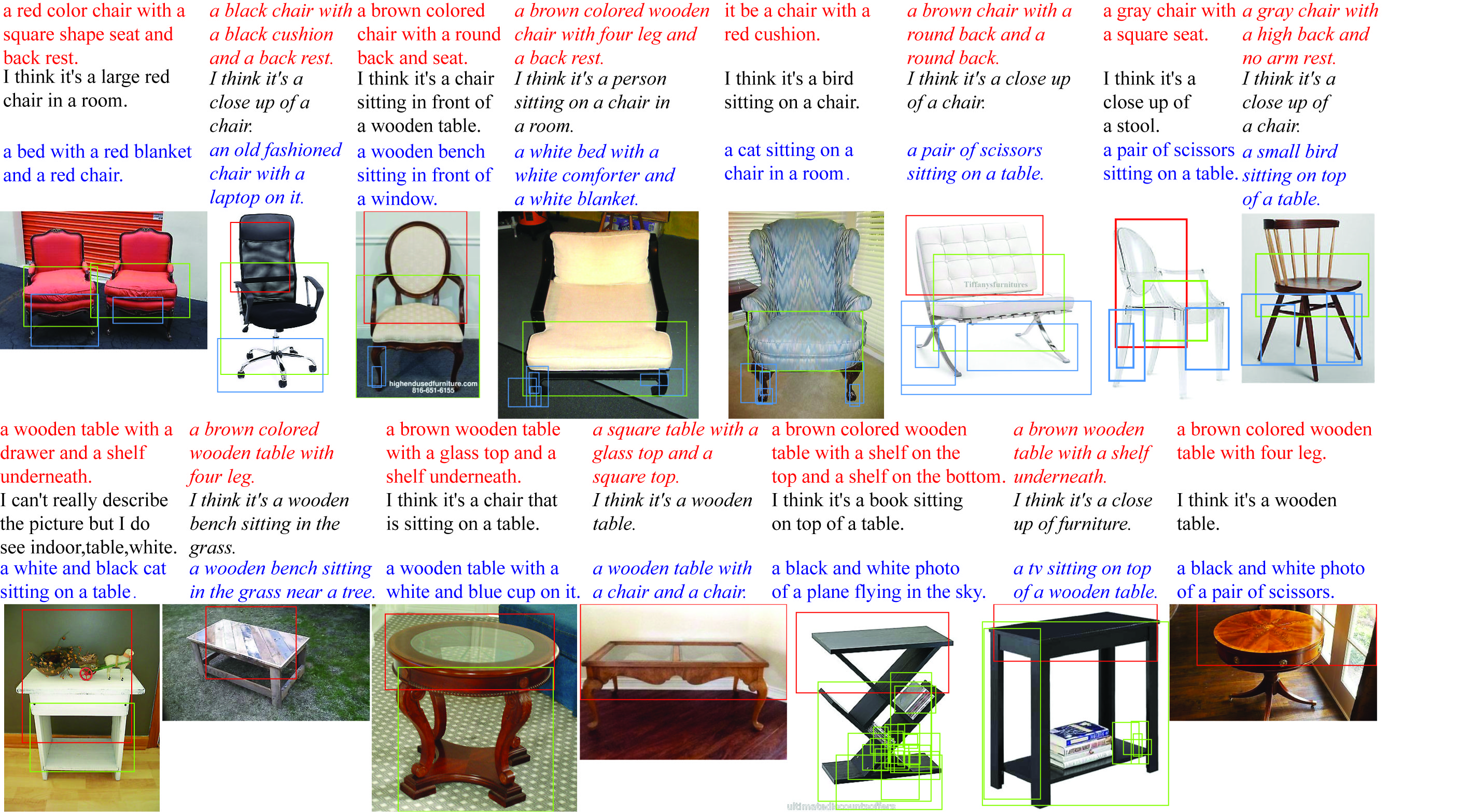}
  %
  %
\caption{\label{fig:CapCompareReal} Compared to CaptionBot (Black)~\cite{NIPS2018_7881} and NeuralTalk2 (Blue)~\cite{Karpathy:2017}, our results (red) presents more detailed and accurate part characteristics under real images, where $\rho>0.5$. The same color in the chair class or table class indicates the same part class. }
\end{figure*}

\noindent\textbf{Comparison with other methods. }We evaluate ShapeCaptioner by comparing it with the state-of-the-art methods. As shown in Table~\ref{table:comp}, we compare to the methods which are able to generate captions from images or image sequences, such as SandT~\cite{Vinyals43274} for image captioning, V2T~\cite{Venugopalan2015} for video captioning, GIF2T~\cite{YaleSongGIF2018} for GIF understanding, SLR~\cite{xushen2018aaai} for video understanding, and $\rm Y^2$Seq2Seq (``Y2S'')~\cite{Zhizhong2019seq} for 3D shape understanding. The results of SandT are produced with mean pooling of all views as shape feature for caption generation, while the results of GIF2T and SLR are produced by the nearest retrieval in the joint feature space of shape and caption.

\begin{table}
  \caption{The comparison with others. $H=32$, $\rho=0.8$, $V=12$.}
  \label{table:comp}
  \centering
  \resizebox{0.5\textwidth}{!}{
  \begin{tabular}{c|c|c|c|c|c|c|c}
    \hline
    Method&B-1&B-2&B-3&B-4&M&R&C\\
    \hline
    SandT~\cite{Vinyals43274}&0.494&0.338&0.251&0.214&0.209&0.381&0.301\\
    V2T~\cite{Venugopalan2015}&0.670&0.430&0.260&0.150&0.210&0.450&0.270\\
    GIF2T~\cite{YaleSongGIF2018}&0.610&0.350&0.210&0.120&0.160&0.360&0.140\\
    SLR~\cite{xushen2018aaai}&0.400&0.170&0.080&0.040&0.110&0.240&0.050\\
    Y2S~\cite{Zhizhong2019seq}&0.800&0.650&0.540&0.460&0.300&0.560&0.720\\
    \hline
    Ours(C)&\textbf{0.937}&\textbf{0.917}&\textbf{0.894}&\textbf{0.878}&\textbf{0.550}&\textbf{0.847}&\textbf{1.789}\\
    Ours(T)&\textbf{0.860}&\textbf{0.755}&\textbf{0.675}&\textbf{0.620}&\textbf{0.362}&\textbf{0.664}&\textbf{1.099}\\
    Ours&\textbf{0.899}&\textbf{0.836}&\textbf{0.785}&\textbf{0.749}&\textbf{0.456}&\textbf{0.756}&\textbf{1.444}\\
    \hline
  \end{tabular}}
\end{table}

We can see that ShapeCaptioner (``Ours'') significantly outperforms the other view-based methods in all metrics, where ``Our'' is the average of results under chair (``Ours(C)'') and table (``Ours(T)'') classes. We believe our results benefit from the ability of understanding parts of 3D shapes, which captures more part characteristics to generate better captions in a more similar way to humans. We also obtain the captions in the test set generated by $\rm Y^2$Seq2Seq from the author, and compare it and SandT (``VMean'') to ShapeCaptioner in Fig.~\ref{fig:CapCompareVis} in all metrics, where we also observe significant improvement.

\begin{figure}[tb]
  \centering
   \includegraphics[width=\linewidth]{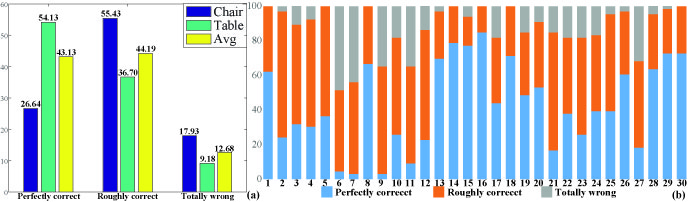}
  %
  %
\caption{\label{fig:User} (a) The statistical analysis of the user study over all cases. (b) The statistical analysis of each case.}
\end{figure}

\noindent\textbf{Real image test. }We further evaluate ShapeCaptioner under a real image set in Fig.~\ref{fig:CapCompareReal}. What we want to show here is that ShapeCaptioner can also help to caption images by leveraging the detected parts, although it is only trained under 3D data. We select real chair and table images from the Stanford Online Products dataset~\cite{SongXJS16}, and we use the trained ShapeCaptioner (``Ours(C)'' and ``Ours(T)'') in Table~\ref{table:comp} to conduct the results. We first compare our method to the state-of-the-art image captioning methods including CaptionBot~\cite{NIPS2018_7881} and NeuralTalk2~\cite{Karpathy:2017} in Fig.~\ref{fig:CapCompareReal}. We find ShapeCaptioner can generate more detailed and accurate descriptions for parts in captions. This benefits from the detected semantic parts with part details, which also justifies that ShapeCaptioner has the ability to overcome the gap between rendered views and real images. Note that the generated captions lack variety due to the absence of training on the real images and captions.

To further evaluate the generated captions, we conduct a user study over randomly selected 30 chairs and tables. We provide all 66 participants each one of the 30 cases accompanied with the generated captions. Then, we ask the participants to select one of three options for each case, i.e., perfectly correct (All descriptions about parts are correct), roughly correct (Some descriptions about parts are correct), and totally wrong (All descriptions about parts are wrong), to evaluate how well the generated caption matches the shape in the real image. Finally, we show the statistical results in Fig.~\ref{fig:User} (a). We can see that participants gave almost $90\%$ cases (``Avg'') perfectly and roughly correct, where the results on each case are elaborated in Fig.~\ref{fig:User} (b). In addition, our results in the table class (``Table'') are better than the results in the chair class (``Chair'') in terms of percentage of perfectly correct. This is because tables are usually simpler than chairs, and there are more training samples in the table class than the chair class in the 3D-Text dataset.

\section{Conclusion}
We propose ShapeCaptioner to better caption a 3D shape by leveraging more part characteristics. ShapeCaptioner successfully learns the ability of semantic part detection from multiple views of 3D shapes in segmentation benchmarks, and effectively transfers this ability to 3D-Text dataset to caption 3D shapes from the parts detected in multiple colored views. Moreover, our part class specific aggregation can also preserve the part characteristics from different views. Our outperforming results indicate that captioning from parts can produce more accurate descriptions for parts, which is also more similar to human's way of describing 3D shapes.

{\small
\bibliographystyle{ieee}
\bibliography{paper}
}

\end{document}